# Video Pre-trained Transformer: A Multimodal Mixture of Pre-trained Experts

**Kastan Day\*, Daniel Christl[×], Rohan Salvi\*, Pranav Sriram[×]**

\*Center for AI Innovation, National Center for Supercomputing Applications, UIUC. [×]Grainger College of Engineering, University of Illinois at Urbana-Champaign

kastanday.com/vpt

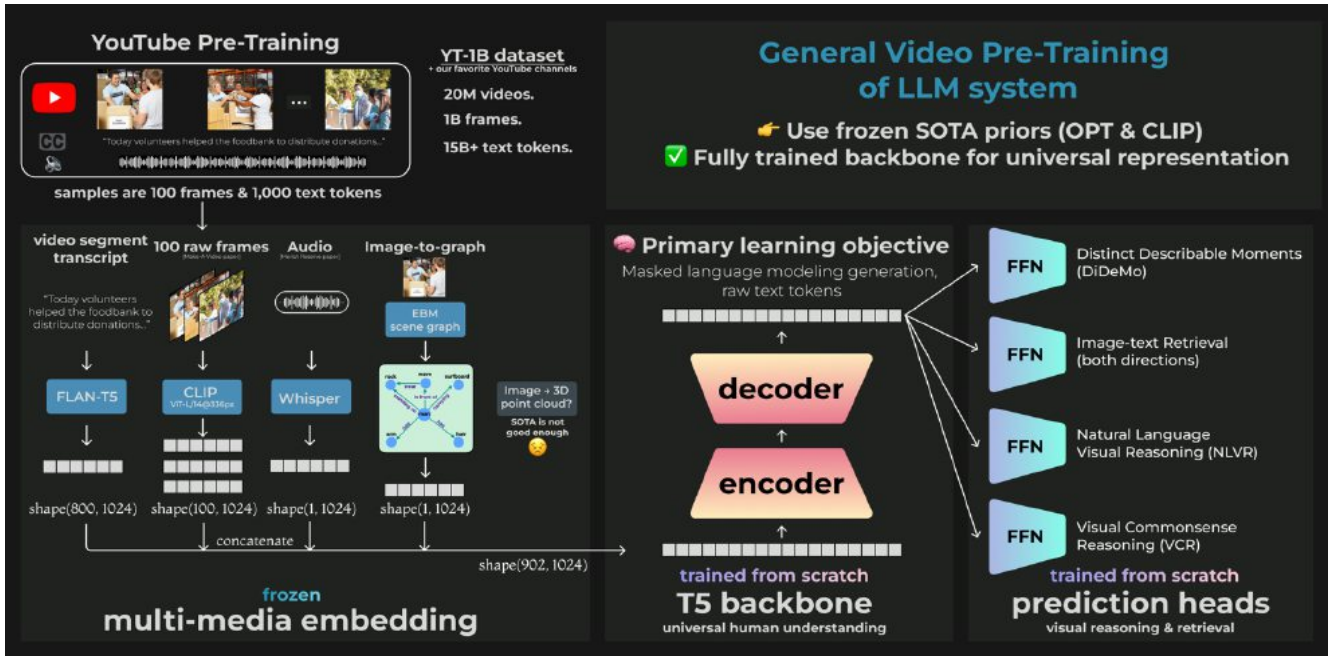

Figure 1: VPT learns multimodal representations of videos from four sources: image-sequences, raw audio, Whisper-generated text captions, and OpenPSG scene graphs). Our backbone is trained to predict the words spoken in a video given encodings of the frames, audio, and scene graph, as well as an encoding of the directly preceding text.

## Abstract

We present Video Pre-trained Transformer. VPT uses four SOTA encoder models from prior work to convert a video into a sequence of compact embeddings. Our backbone, based on a reference Flan-T5-11B architecture, learns a universal representation of the video that is a non-linear sum of the encoder models. It learns using an autoregressive causal language modeling loss by predicting the words spoken in YouTube videos. Finally, we evaluate on standard downstream benchmarks by training fully connected prediction heads for each task. To the best of our knowledge, this is the first use of multiple frozen SOTA models as encoders in an "embedding → backbone → prediction head" design pattern – all others have trained their own joint encoder models. Additionally, we include more modalities than the current SOTA, Merlot Reserve, by adding explicit Scene Graph information. For these two reasons, we believe it could combine the world's best open-source models to achieve SOTA performance. Initial experiments demonstrate the model is learning appropriately, but more experimentation and compute is necessary, and already in progress, to realize our loftier goals. Alongside this work, we build on the YT-20M dataset, reproducing it and adding 25,000 personally selected YouTube videos to its corpus. All code and model checkpoints are open sourced under a standard MIT license.

# 1 Introduction

Humans receive a variety of stimuli to help interpret the surrounding environment: vision, audio, smell, etc. and jointly use these signals to understand the world. These events *inform* each other and compound understanding in ways that individual modalities are incapable of producing. Machines, conversely, are often limited to maximize performance on individual modalities. Many SOTA models excel in one or two modalities. Then, what might occur when combining these models together, having each inform the other to complete a more coherent scene?

It is with these ideals in mind that we present VPT, a joint transformer trained from scratch to combine different modalities. This model takes advantage of transfer learning, with embeddings created from multi-billion parameter SOTA models CLIP for images (Radford et al., (2020)), Whisper for the highest quality YouTube captions including word-level timings (instead of YouTube's provided captions and phrase-level timings) (Radford et al., (2022)) and Open-PSG for explicit scene graph relationships, based on Detectron2 image segmentation (Yang et al., (2022)).

We plan to train this model at scale, using a dataset with over 20 million videos in addition to many handpicked videos that epitomize visual-audio relationships. Currently, we train on only 25k videos. For loss, we implement self-supervised learning across the captions and finetune on the VQA dataset.

The end goal of VPT is general knowledge based on video input, producing parameters that can be used as input to a variety of heads for downstream tasks. In summary, our key contributions are the following:

a) We introduce the VPT architecture which leverages existing SOTA models, one or more per input modality, to produce embeddings which are integrated in a large universal backbone transformer.
b) In recognition of the power of open source ML, such as the innovation enabled by the release of BERT and Stable Diffusion, and in protest against closed-source models from OpenAI and Microsoft Research, the final PyTorch models will be available under a standard open source MIT license and will be trivially consumable via Huggingface.
c) Including explicit scene-graph information in a multi-modal video model.
d) Employing Whisper for the purpose of generating millions of captions, a novel large-scale usage of Whisper for training a transformer with caption-frame-scene graph tuples.

Overall, our work demonstrates the potential power of utilizing data in structured and unstructured ways and speaks to the strength of combining multiple modalities.

# 2 Related Work

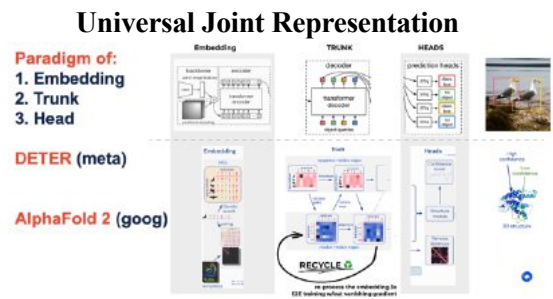

Figure 2: An architecture diagram of the Embedding, Trunk, Head design pattern as applied in Carion et al., (2020) and Jumper et al., (2021).

A variety of world-class self-supervised models have been introduced with the same design pattern: Embeddings → Backbone → Prediction-Heads, which is the central inspiration for this work (Carion et al., (2020), Caron et al., (2021), Jumper et al., (2021), Li et al., (2022). , Tesla AI Day (2022).)

**Universal Transformer from Google Brain (**Liu et al., (2020)**)** introduced using a transformer layer to combine domain-specific encodings. The authors extracted the last hidden layer from a set of task-specific CNNs as input to a transformer layer to create a "universal representation" for diverse downstream image classification. This concept inspired our work.

**Merlot and follow-up Merlot Reserve from the Allen Institute for AI (**Zellers et al., (2021)**,** Zellers et al., (2022)**)** introduced an open-source model for multimodal video understanding achieving SOTA performance in a wide range of downstream tasks, in both zero-shot and fine-tuned settings. Moreover, the same work introduced the largest curated YouTube dataset to date, in the form of YouTube video IDs/URLs, on which we build for this work.

The authors used standard contrastive frame-transcript matching, much like CLIP (Radford et al., (2020)), but also introduced a novel learning objective for Masked Language Modeling (MLM) with attention-guided attention. The problem with normal/naïve MLM on YouTube data is that random masking often masks uninformative filler words like "umm" and "you know." Therefore, Merlot prioritizes masking tokens which are most attended to by the joint vision-language encoder to mask the most helpful, informative, and visually grounded tokens to provide a richer signal to MLM learning. The authors use a 20% masking ratio, and 50% of the time masks a random word and 50% of the time masks one of the top 20% most-attended-to tokens. In follow-up work, Merlot Reserve added audio to their pre-training objective via a contrastive audio masking objective, much the same as MLM. Ablations show audio provided additional information not amenable to transcripts, and increase performance on Situated Reasoning by 1.1%.

**SimCLR from Google Brain, Toronto (**Chen et al., (2020)**)** introduced L2-normalized embeddings for contrastive learning, as used in this work.

### The more loss terms the better in SSL

**FLAVA from FAIR (**Singh et al., (2022)**)** utilized a large collection of loss functions. During pre-training, they employed Masked Image Modeling (i.e., hidden image patches), MLM, contrastive masked multimodal modeling (MMM) and contrastive image-text matching (ITM). The design is very similar to Merlot RESERVE. Finally, like this work, FLAVA then learns task-specific classification heads from random initializations for evaluations on VQA, GLUE and ImageNet.

**AlphaFold2 from Google Deepmind, London** demonstrated the success of using many loss functions, seven in their final model, especially when each is carefully tailored to performance on downstream tasks. Their design can be approximated with human expertise and should be confirmed via extensive ablations testing all combinations and variants of the loss functions.

### Creative Application of Attention

This approach does not just combine transformers. Rather, it creatively applies attention to filter and combine vectors. (Ma et al., (2022), Jaegle et al., (2021), Jaegle et al., (2022), Bertasiu et al., (2021))

**Perceiver and follow-up Perceiver IO, DeepMind, London** (Jaegle et al., (2021), Jaegle et al., (2022)) Solving a similar problem as LongFormer, Perceiver uses cross attention in place of less self-attention over the input dimension to enable $O(N)$ linear compute scaling on input sequences, instead of standard $O(N^2)$ self-attention. Their follow-up, PercieverIO, extends this efficiency to the decoder as well and enables something akin to a standard autoencoder but with very large (audio, video, and label) inputs and outputs. The result is an impressive reconstruction of the original video, audio and a class label prediction. Nevertheless, their images are small; each frame is only 512 pixels, much less than the standard 224x224 = 50k pixels used in CLIP and X-CLIP and even smaller than ViT's 32x32=1,024 pixel patches. In our work we don't see sufficient benefit in attempting to reconstruct full video outputs, as is so expensive in PerceiverIO, and instead only predict textual outputs.

**X-CLIP by Microsoft Research Asia, Xiamen University China (**Ma et al., (2022)**)** greatly informed how we think about temporally aligning the words someone speaks with the actions appearing on screen; in short, often YouTubers don't speak and act simultaneously. They typically talk first, then show. This underscores the importance of having video-level features and not just one-frame per one-caption. X-CLIP also leveraged Contrastive learning at every combination of temporal lengths i.e. all combinations of {video, frame, sentence, word}.

**CLIP-Event by UIUC (**Li et al., (2022)**)** optimized SSL with a "hard negative" sampling strategy to ensure negative samples were as challenging as possible, i.e., events with similar visual features but different labels. This inspires our future work of retrieval augmented hard negative sampling, where negatives are not selected randomly from the training batch, but rather are retrieved from any part of the training corpus.

**Chinchilla from Google Deepmind (**Hoffmann et al., (2022)**)** introduced the "updated" scaling laws and recommended the use of 20 text tokens per model parameters during pre-training (e.g., 1B-param models should use 20B tokens in dataset for compute-optimal performance). This implies the first generation of "breakthrough LLMs" (e.g., GPT-3, OPT, T5, PaLM) were over-sized for the size of their pre-training data, if, that is, we want to be "compute optimal" in some sense. Chinchilla also inspired the style of this background section.

## 3 Method

We present VPT, a causal language model which accepts multimodal embeddings (generated by state-of-the-art frameworks) as input and autoregressively predicts the transcripts of YouTube videos as output. In addition to our proposed model, we present an elaborate data preparation procedure to construct the input embeddings for the pre-training process.

We begin by downloading 28 million videos from YouTube. For each video, we run a state-of-the-art speech recognition system Whisper (Radford et al., (2022)) to generate a transcript with per-word timestamps and other metadata. In a similar fashion to previous works (Zellers et al., (2021)), we employ a sliding window technique to partition the video into word-dense segments containing 15 words each. We extract one frame per video segment sampled at regular intervals to use as our visual modality. We feed these frames into a SOTA scene graph model OpenPSG on this image to generate a context-descriptive scene graph. Thus, we have three resulting modalities: image, text, and scene graph. We embed all the modality features using CLIP (Radford et al., (2020)) to generate an input embedding of size [3, 768], which constitutes one training example.

### 3.1 Model Architecture

**Whisper for SOTA caption quality (**Radford et al., (2022)). Spoken word is a heavily underutilized source of data in large language models. This is partially due to their noisy nature and the difficulty in producing accurate transcriptions. However, similarly to Merlot (Zellers et al., (2021)), we can still leverage the additional information in spoken language from videos indirectly via caption generation. While Merlot leverages YouTube's auto-generated captions, we attempt to improve upon model performance by utilizing Whisper, a SOTA ASR model for more accurate caption extraction.

Whisper is a large speech recognition model that generates transcripts from audio files. It applies zero-shot transfer from the original training to a new environment without the need for finetuning. This model served three major functions for our project. Foremost, it auto-detects language, making it simple to remove non-English videos from our dataset. Second, it boasts impressive accuracy, making it better for word-to-text translation than the auto-generated captions from YouTube. Finally, Whisper, when used with library Lhotse, can extract a high precision timestamp for each individual word. This feature enables further refining of the preprocessed by ensuring the chosen frame is a known distance from the spoken words, and ensuring we use videos and video-segments with sufficient word density. We only keep video segments with meeting a threshold of 30 words per minute speech.

**Scene graph for image contextualization.** While images alone are a powerful source of information, their lack of structure and contextualization can lead to mediocre performance over a short duration of training. Thus, we propose an additional source of information acquisition from these frames: scene graphs.

**Visual Genome (**Krishna et al., (2016)) define scene graphs as a structured formal graphical representation of an image. Scene graphs represent a web of relationships among interconnected objects and have led to the development of many powerful models in image captioning, image retrieval, visual question answering, relationship modeling and image generation.

In our work, we employ OpenPSG, (Yang et al., (2022)) a state-of-the-art scene graph generator with Panoptic Scene Graph generation (PSG). Existing scene graph generation tasks (Tang et al., (2018)) in (a) use bounding box-based labels, which are often inaccurate — pixels covered by a bounding box do not necessarily belong to the annotated class — and cannot fully capture the background information. In contrast, OpenPSG leverages PSG in (b) to construct a more comprehensive and clean scene graph representation. This representation contains more accurate localization of objects and description of relationships with the background i.e., the trees, pavement, and sky. We hypothesize that this will enhance the innate scene classification and spatial intelligence of our model.

|  | Accuracy | Iterations |
|---|---|---|
| Base T5 + VQA Finetuning | 0.2390 | 20k |
| Base T5 + YouTube Pre-training + VQA Finetuning | 0.2389 | 5k |
| Base T5 + YouTube Pre-training + VQA Finetuning | 0.2468 | 10k |
| Base T5 + YouTube Pre-training + VQA Finetuning | 0.2555 | 20k |
| Base T5 + YouTube Pre-training (only Yes/No) | 0.2389 | 20k |
| **Base T5 + YouTube Pre-training + VQA Finetuning (w/o scene graph)** | **0.3098** | 20k |

Figure 3: Results on VQA benchmark. We show that YouTube pre-training improves downstream performance on VQA. We also conduct ablations against including scene-graph information and are surprised to find that, as implemented now, scene-graphs hurt performance.

**CLIP for image and text encoding.** While other models (Zellers et al., (2021)) train their own image and text encoders, we hypothesize that a self-supervised representation learner such as CLIP may enhance generalizability to downstream tasks due to its demonstrated zero-shot capabilities (Radford et al., (2020)). As CLIP is trained in a contrastive manner on a large dataset, its generated embeddings are task-agnostic and should be sufficiently powerful to enable the effective pre-training of our model. Moreover, using a frozen, pre-trained encoder greatly reduces the size and number of trainable parameters of our model, increasing the portability and simplicity of our model.

We use the most capable pre-trained CLIP model (VIT-L/14@336px) and preprocess our entire YouTube dataset with both the CLIP text and image encoders. For a given segment of a video, we encode the Whisper-generated caption of the segment with our CLIP text encoder and encode some number of evenly sampled frames (a tunable hyperparameter) from the segment with the CLIP image encoder. These embeddings are then saved and used for the primary training with our T5 backbone.

### 3.2 Supervised Training

Our model employs auto-regressive causal language modeling (Chung et al., (2022)) based on the generated captions. Specifically, we explore two different options for utilizing captions as ground truth labels: our first method is to take the entire caption embedding as input and use the raw caption text as output. In this setting, we hope that the model can quickly learn to combine the input embeddings in a way that encapsulates the scene's key information. Our second method is to encode the first half of the 15-word caption and have the model predict the second half the raw caption text. In this setting, we hope that our model also learns to jointly combine the input embeddings albeit more slowly. In this alternative, there is no leaky data.

## 4 Experiments

### 4.1 Visual Question Answering (VQA)

We evaluated our model through the VQA dataset (Agrawal et al., (2015)) to examine its vision, language and commonsense understanding. The VQA is a benchmark visual question answering dataset consisting of 265,016 images and over 1 million open-ended questions.

**VQA Task.** A model is given an image (from MS COCO dataset) and an open-ended question. Additionally, 5 captions associated with the image can be used as input, too. The model needs to generate a natural language response (90% of time the correct answer is a single word). The output generated by the model is compared with 10 ground truth answers and scores as per an evaluation metric provided by the authors of the dataset. We evaluate our performance using the standard VQA accuracy metric, described by the following formula:

$$\text{Acc}(ans) = \min\left\{\frac{\#\text{humans that said } ans}{3}, 1\right\}$$

**Finetuning Approach.** We use the VQA v2 dataset (Goyal et al., (2016)) for finetuning and evaluation. Specifically, we finetune on the training dataset for 1 epoch with a batch size of 16 and an AdamW optimizer with learning rate 0.0001. For a given image-question pair, we extract the scene graph from the image using OpenPSG and the CLIP embeddings for both the question and the image. We follow a near-identical concatenation procedure to our pre-training setup, except we replace the video caption embedding with our question embedding. As there are multiple possible answers provided for a given sample question, we randomly pick one of the provided answers and use it as our ground truth annotation for the visual question. We then train the model with teacher forcing to output the expected answer (truncated/padded to a fixed sequence size of 128) and use regular cross-entropy as our loss function.

We train the T5 base model variants for (and up to) 20k iterations on the training dataset. We visualize the loss curve and corresponding accuracies of a specific variant at iteration 5k, 10k and 20k in the results table. While the improvements are minor, the model still sees increases in accuracy throughout all 20k iterations, a sanity check to ensure we are not overfitting.

**Ablations.** We also run ablations to further examine the contributions of the large-scale pre-training and our novel scene-graph incorporation. We first examine the effectiveness of the YouTube pre-training quantitatively on the VQA task. There is an accuracy improvement of 1.65% with the YouTube pre-training, which is quite insignificant considering the scale and compute required for this effort. This may be due to an overarching issue regarding the training objective, which is susceptible to abysmal performance as a result of information leakage. In the pre-training task, our training objective is to reconstruct the CLIP-encoded text caption with our T5 model. However, recall that this text caption is fed in as an input to the model; this training objective may be too naïve and weak to propel the learning of strong representations of downstream tasks.

We also examine the contributions of the scene-graph modality to the downstream VQA performance. Contrary to our intuitive hypothesis regarding the utility of scene graphs, the model seems to perform considerably better without the scene graph modality on the VQA task (an accuracy improvement of ~6-7%). While this finding does not invalidate our hypothesis about the utility of the scene graph modality, it indicates that our usage of the scene graph information is inappropriate. Currently, we embed the scene graph with the CLIP text encoder and simply concatenate it onto our input embeddings. However, the scene graph contains multiple phrases that focus on different semantic portions of the image. Encoding all these phrases simultaneously with the CLIP text encoder is likely an issue. Moreover, scene graph information is structured and could be used more intelligently to improve our learned representations (as opposed to simple concatenation). An initial idea is to use scene graph information to refine image embeddings with a cross-attention mechanism, thereby leveraging it in an auxiliary manner rather than an explicit modality.

Finally, in response to our extremely general low performance on the VQA task, we ran an ablation to simplify the task by running training and evaluation on only yes/no questions in the dataset. The experiment would also enable us to reason about possible issues that arose due to the teacher forcing framework. This ablation showed similar results, again accentuating a training collapse perhaps due to larger issues with our proposed framework. We investigate these issues further in the qualitative analysis section.

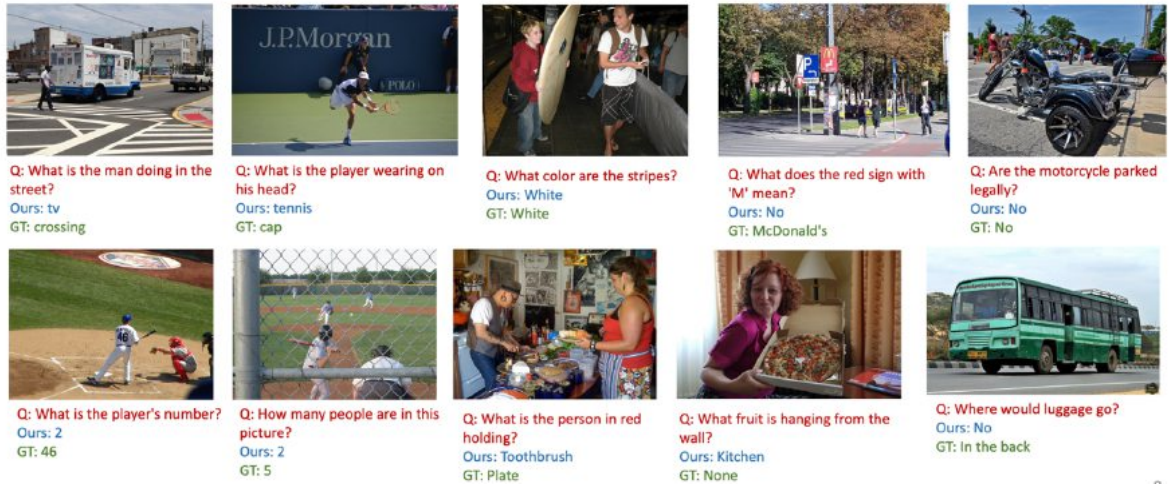

Figure 4: Qualitative analysis of model performance. Even incorrect examples reasonable guesses, but the model often collapses to predicting "no" unnecessarily.

## 5 Qualitative Analysis

We visualize our model outputs on select image-question pairs to qualitatively assess the model and reason about the low accuracies mentioned in the previous section. Through the visualization of our outputs (many beyond those visualized above), we noticed that the low accuracies of our model variants are due to a collapse in the model predictions to a trivial set of outputs. For example, the model simply predicts "No" for a majority of the questions (regardless of the context) interspersed with some predictions of "0" as well. All the variants except the scene graph ablation were tremendously susceptible to this collapse, as we struggled to find even a small number of samples with "non-trivial" predictions. The scene graph ablation exhibited slightly more promising behavior (albeit largely similar), which we visualize in Fig. 4. The first three images in the top row illustrate that our model possesses some understanding of the question and the semantic context of the image. However, the model regresses to nonsensical and sometimes trivial in most of the other images, demonstrated by the prediction of "No" and "2" in certain cases.

Our qualitative results corroborate our quantitative findings and our reasoning behind the potential sources of the general low accuracy across all variants. The observed trivial set of outputs could likely be due to a combination of overfitting along with a poor choice of training objective and model architecture, as well as other design choices e.g., model input structure.

## 6 Conclusion, Limitations, Future Work

### 6.1 Conclusions

The goal of VPT is to provide general knowledge based on video input. This is achieved by leveraging multiple multimillion parameter models to create embeddings for a large-scale joint transformer via transfer learning. Scene graphs are also created for each frame to provide additional context to the embedding space. The model is trained at scale using a dataset of over 20 million videos, and self-supervised learning is used to optimize the model. The end result is a model that fuses multiple modalities (including a novel structured scene graph modality) to enable universal representation learning, enhancing performance on myriad downstream tasks.

### 6.2 Limitations and Future Work

**Training Data Usage.** In our current model, we are training on only a fraction of available preprocessed data. Scale is one key factor in large deep learning models, (Brown et al., (2020)) and we believe that with more compute time/power our model will better acclimate to its input embeddings and better capture information in its output embeddings.

**Loss Function.** In our current loss function, we are using the full caption as both input data and as the output labels. This means our model contains information about a variable that is being predicted by our model (leaky data). This likely led to overfitting, culminating in poor performance on our downstream task. To rectify this issue, we propose using our alternative method of splicing the caption in half as mentioned in the methods section.

Additionally, our model will ideally leverage many more loss functions in future iterations. Options we're currently implementing include contrastive image-caption matching, position-encoding based temporal frame ordering, and contrastive caption-audio matching. Based on previous work, (Zellers et al., (2021), Zellers et al., (2022)) these objectives could encourage the model to learn more robust representations.

**Audio**. One more major modification to our model is the inclusion of audio embeddings. Raw audio is capable of encapsulating information that language cannot, like tone and surrounding sound. This is particularly relevant in video. Proper utilization of audio embeddings has been shown as a source of improvement in other multimodal transformers (Zellers et al., (2022)). Additionally, we plan to leverage a pre-trained audio transformer (Gong et al., (2021), Gong et al., (2021), Verma et al., (2021) ) as opposed to a self-trained transformer, relating to our overarching theme of transfer learning. Additionally, our model might be able to leverage different audio embedding models with different downstream goals. One instance of this could be using one embedding model trained for tone recognition and another embedding model trained for sound recognition.

## Acknowledgements

Heng Ji, Yuxiong Wang, and the NCSA Delta admin team.

**Appendix Contents**

We provide the following materials:
- A description of our training dataset and recommendations for multimodal data management in Python (Section A)
- A discussion on instruction finetuning (Section B)
- Details of our compute usage (Section C)
- Limitations (Section D)
- Details about why we chose a T5 backbone (Section E)
- Future Datasets/Downstream Tasks (Section F)
- Supplementary Materials (Section G)

## A  Pre-training Dataset

Our dataset is based on YT-1B (Zellers et al., (2022)) with additional handpicked videos. Introduced in Merlot Reserve from The Allen Institute for AI, YT-1B contains the URLs of 20 million public YouTube videos, or 1B frames if you take one frame per 12 seconds. The authors filtered it to be "visually grounded" with audio & images that are strongly related, they provide details in their appendix. It is an evolution of their prior work *Merlot*, which introduced YT-Temporal-180M. ([download link](#)) This dataset was created by first collecting a diverse collection of 200 million videos via breadth first search through recommended videos from a variety of starting videos. This gathered 2 million unique channels, a metric that demonstrates diversity in the video choices. Next, videos with non-English titles were removed. Then, an original language model was trained with the goal of maintaining visually-grounded videos. This model was trained via predicting metadata of videos based on manual annotations on 2000 different videos. This process narrowed the group of videos down to 30 million. Finally, videos with overly-dense spoken work were filtered out, along with videos whose thumbnails were too similar to other videos already in the dataset. With this process complete, 20 million high-quality videos remained.

In addition, since this team is personally passionate about YouTube, we added hand-selected channels that typically have real-world visual action, accompanied by voiceover. Voiceover is great because typically the transcript is temporally aligned with the action on screen, i.e. YouTubers show B-Roll of the topic they're describing. This is in contrast to unscripted content YouTubers, whom often describe what they're *going to* do long before they show doing it on camera. This temporal misalignment poses major challenges for learning joint representations, as explored in depth in X-CLIP (Ma et al., (2022)).

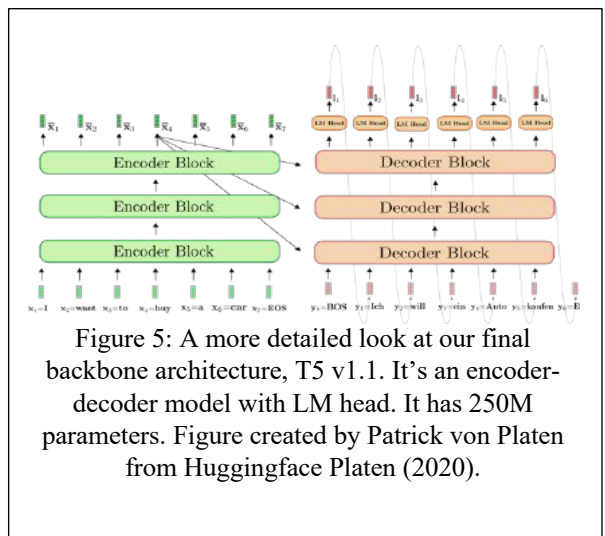

Figure 5: A more detailed look at our final backbone architecture, T5 v1.1. It's an encoder-decoder model with LM head. It has 250M parameters. Figure created by Patrick von Platen from Huggingface Platen (2020).

For our handpicked process, we manually determined channels wherein the spoken word had high relevance to the visuals presented in the video. Then, we downloaded the entire channel's worth of content. To assist in generating potential candidates we utilized the GPT-3 playground with prompts asking for vlog or otherwise highly visual channels. With these suggestions, we manually reviewed the channels to confirm whether the visual-audio relevancy was high. This process accounts for 23,582 additional videos to our dataset. The following list represents channels that we incorporated into our dataset.

- athleanx
- atomicfrontier
- berm_peak
- cinemasins
- corridor_crew
- johnnyharris
- kurzgesagt
- matt_davella
- minutephysics
- mythbusters (a collection of visually grounded clips, not full episodes)
- nilered
- planetearth
- realengineering
- theactionlab

- tomscott
- veritasium
- wall_street_journal
- wendover

**Dataset storage**

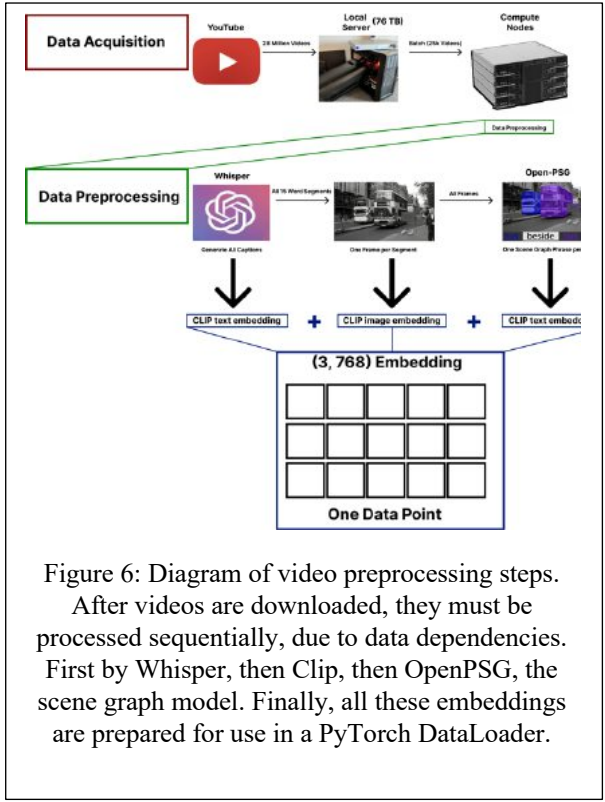

Figure 6: Diagram of video preprocessing steps. After videos are downloaded, they must be processed sequentially, due to data dependencies. First by Whisper, then Clip, then OpenPSG, the scene graph model. Finally, all these embeddings are prepared for use in a PyTorch DataLoader.

The 2.8 million videos collected from YouTube so far took 6 weeks to collect on a symmetrical 1-gigabit internet connection. It is 330 thousand hours (38 years) of video and requires 76 TB of storage. It's downloaded in 360p, and although 240p has sufficient resolution, the compression applied by YouTube is so aggressive that the 360p videos have much greater visual fidelity. We downscale the videos using OpenCV.

## B  Extending *instruction finetuning* to multimodal models

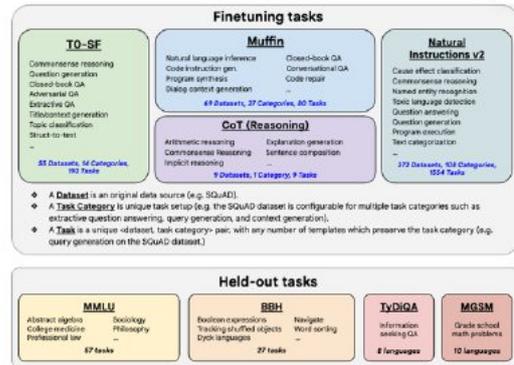

Figure 7: Overview of the benchmarks Flan-T5 used as supervised fine-tuning data to achieve its SOTA performance (Chung et al., (2022))

Flan-T5, one of the best open source GPT-3 competitors, was pre-trained on Common Crawl (C4) and fine-tuned on a gargantuan 473 datasets. This amounts to using internet-scale data for pre-training, every possible ML and linguistics benchmark as fine-tuning data. The authors are using every imaginable data-source.

Similarly, for our conference-quality follow-up we intend to fine-tune on video benchmarks:

| Task | High-quality Datasets |
|---|---|
| 🔑 Key dataset for generic pre-training of Trunk Transformer. | **YT-1B + hand-picked YouTube channels and TV shows** |
| Unsupervised video pre-training. | |
| Image-text Pre-training | COCO, VisualGenome, SBU ConceptualCaptions |
| Image-text Retrieval (both directions) | COCO, Flickr30k |
| Visual Question Answering | VQAv2, OKVQA, A-OKVQA |
| Natural Language Visual Reasoning (NLVR) | NLVR2 |
| Visual Entailment (VE) | SNLI-VE |
| Video-text Retrieval (both directions) | MSRVTT, DiDeMo, MSVD, LSMDC, Activity Net |
| Video Dialogue | AVSD |

| Task | High-quality Datasets |
|---|---|
| Non-standard | EgoCentric 4D |

Figure 8: Overview of the benchmarks we intend to use for both evaluation and, subsequently, for fine-tuning of our SOTA contender.

## C  Compute Size

Performing inference with our preprocessor models (Whisper, CLIP, OpenPSG, and, eventually, an audio transformer, and Flan-T5 text encoder) requires significant computation and engineering effort. It requires 4 GPU-years to run Whisper-base on 1M videos on a HPC system (4xA40 GPU, 64-core EPYC 7763 CPU and 256 GB RAM). Significant engineering graft was devoted to achieving an average of 94% GPU utilization, which is far above industry standard for LLM workloads.

CLIP had similar compute requirements, but was about 15% faster than Whisper. This is largely because we only use 1 frame per 15-word video segment, meaning there is less overall visual data to process. OpenPSG ran significantly faster due to running on top of a highly optimized Detectron2 implementation.

**Data storage**
As we implemented our inference system, we tried saving data in nearly every form of data storage in Python: np & pd → Jsonlines → Parquet, PyArrow → Dask, Zarr, Xarray → NetCDF4, HDF5. Here are a few recommendations.
For text data, I would highly recommend PyArrow. In the end, we used mostly NumPy and JsonLines for this paper, but are now transitioning to an ML-specific database called DeepLake (DeepLake Database). Another few contenders worth investigating are DuckDB (DuckDB) and Lance (Lance Database).

We have strict storage requirements:
- Thread-safe & distributed,
- Columnar O(1) lookups,
- Compressed storage,
- Ragged arrays (i.e., support for various media types)
- Atomic writes (safe for preemptible VMs)

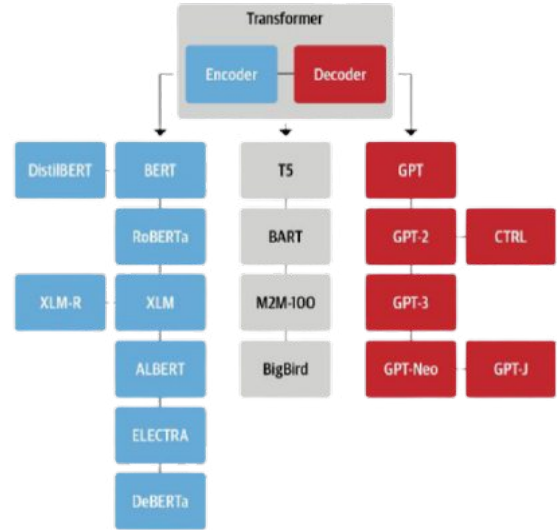

Figure 9: A family tree of "1$^{st}$ generation" LLM Transformer architectures, from the Huggingface textbook. https://transformersbook.com/

- Data streaming & S3 support (nice to have).

## D  Limitations

Currently, all embeddings are created with the CLIP image or text encoder. Originally, we wanted to use text embeddings from Flan-T5 and scene-graph embeddings directly from OpenPSG.

Nevertheless, unilaterally employing CLIP has the advantage of creating an aligned latent space. Since all embeddings are in the same space, we don't have to worry about using contrastive losses to align the embeddings produced by different encoder models. However, CLIP's text model is not ideal for encoding the verbose language used on YouTube because it is not inside CLIP's 'image caption' domain. This simplified implementation at the expense of performance.

## E  Choice of Backbone Transformer

We elected to use a pre-trained model of a standard SOTA architecture for our backbone transformer. Theory and testing are described below.

**LLM Architecture Best Practices**

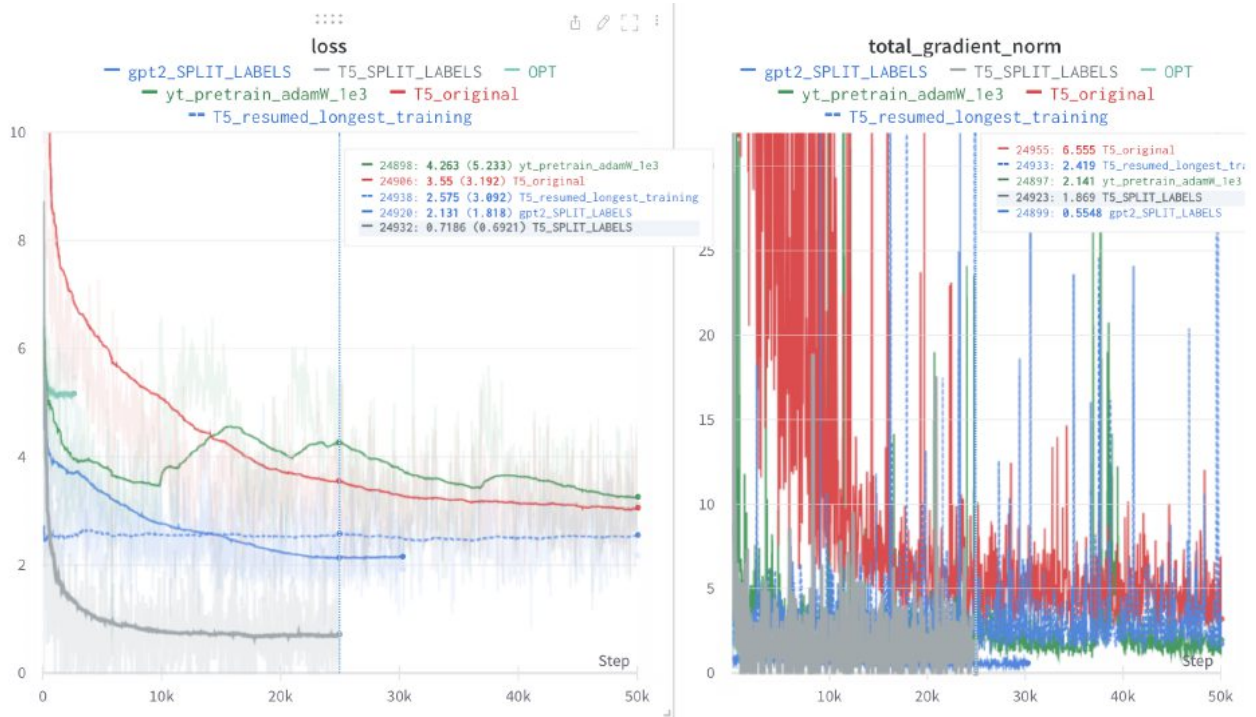

Figure 10: (Left) Loss curves and (right) the norm of the gradient for training VPT ablations on 25,000 input videos. T5 produced the best loss results, followed by GPT-2, then OPT. Furthermore, training improved when only the first half of the labels were used as input, and only the 2nd half being used as targets. This prevents info leaking, thus preventing learning shortcuts.

As shown in Figure 9 there is no standard consensus on the best architecture design of encoder-only vs encoder-decoder vs decoder-only models. However, recent work from Google's PaLM team notes that "to date, most large language models (i.e., more than 100B parameters) are trained as decoder-only casual language models (e.g. GPT-3, Gopher, PaLM). Meanwhile, bidirectional models (e.g., BERT, T5, ST-MoE) have also been popular as the model of choice, especially in smaller computational regimes (e.g., less than 30B parameters and often times in the ranges of hundred (sic.) of millions of parameters) (Tay et al., (2022)). This recommendation is consistent with our results.

In short, combining PaLM's observations with the Updated Scaling Laws proposed in Chinchilla (Hoffmann et al., (2022)), the best practice is to use a decoder-only causal language model architecture if your data size indicates your model should have more than 100 billion parameters. For smaller datasets, elect for encoder-decoder models. Currently, I am unsure of best-practices for encoder-only architectures.

**Ablation of different pre-trained backbone transformers**
Due to ease of implementation, we experimented on these three models. See figure 10 for loss curves.
- T5 v1.1 (`google/t5-v1_1-xxl`)
- T5 (`t5-11b`)
- Flan-T5-XXL (`google/flan-t5-xxl`)
- GPT-2 (`gpt2`)
- OPT (`facebook/opt-13b, 66b`)

**Embedding Dimensions**
We chose these backbone sizes because each had an embedding dimension (d_hidden) of 1024, which was compatible with all of our upstream Expert models. We elected for the largest embedding dimension possible while being supported by all Expert models. We wanted to avoid using an 'adapter' (i.e. 1 or 2 layer learned FCN/MLP) between the embedding produced by CLIP and the embedding dimension of the backbone transformer because we did not want to risk degrading the information in our SOTA embeddings.

## F   Future Evaluation Datasets

**Video Language Event Prediction (VLEP)**
Video-and-Language Event Prediction (VLEP) is a benchmark dataset for fine-grained future event prediction from videos. It contains 28,726 future event prediction examples (along with their rationales) from 10,234 diverse TV Show and YouTube Lifestyle Vlog video clips.

In this dataset, the model is provided with a video clip, aligned dialogue and two future events as input. The model needs to predict of the two events which is most likely to occur. To make such future predictions requires model to understand not only videos and dialogue in depth, but also requires a significant amount of multimodal commonsense knowledge.

**Distinct Describable Moments (DiDeMo)**
Distinct Describable Moments (DiDeMo) is a benchmark dataset for retrieving localized moments in a long video through natural language queries. It contains a total of 26,892 moments and one moment could be associated with descriptions from multiple annotators. The descriptions in DiDeMo dataset are detailed and contain camera movement, temporal transition indicators, and activities.

For this evaluation the model receives a video clip (approx. 30 seconds long) and a natural query. The model then has the task to retrieve the moment from the video that best matches the query. To perform such a video retrieval task the model must possess profound video and language understanding. This benchmark evaluates how well the model captures temporal dependencies and reasoning with respect to the text.

**TV Series Question Answering (TVQA)**
TVQA dataset is another benchmark question answering dataset that consists of 152,545 QA pairs from 21,793 video clips. In this task, the model receives a video clip with 7 questions. Each question has 5 answers (1 correct). The dialogue (character name + subtitle) for each video clip is also provided. The model is required to predict the most likely answer among the five probable answers for each question.

In the TVQA dataset the questions are designed to be compositional in nature: it requires systems to jointly localize relevant moments within a clip, comprehend subtitle-based dialogue, and recognize relevant visual concepts. While some questions can be answered using subtitles or videos alone, some require information from both modalities.

**Visual Commonsense Reasoning (VCR)**
The Visual Commonsense Reasoning (VCR) dataset contains 290k multiple choice questions derived from 110k movie scenes, with a focus on visual commonsense.

## G  Supplementary Material

Follow the project homepage for updates as they happen: http://kastanday.com/vpt

All project code and full git history can be found on GitHub: https://github.com/KastanDay/video-pretrained-transformer

All training logs can be viewed on the Weights and Biases page (new training logs will be published from the GitHub README): https://wandb.ai/kastan/VPT-custom-t5

The models will be made available on the Hugging Face Hub: https://huggingface.co/kastan

## H  Limitations

As hypothesized in our final presentation, we just discovered that our current evaluation strategy on VQA has at least one fundamental problem leading to non-representative performance results (shown in Figure 3). Because we're using Huggingface transformers in ways that were never intended, we re-wrote the training loop of T5 to accept our non-standard inputs. However, we neglected to make the subtle alterations to the inference loop, and now need to re-write the Huggingface `model.generate()` function to support our customized architecture. This work is in progress but requires considerable engineering effort. We will update Heng Ji with new results as they are available.